\pdfoutput=1

\documentclass[11pt]{article}

\usepackage[preprint]{acl}

\usepackage{times}
\usepackage{latexsym}

\usepackage[T1]{fontenc}

\usepackage[utf8]{inputenc}

\usepackage{microtype}

\usepackage{inconsolata}

\usepackage{graphicx}

\usepackage{amsmath}
\usepackage{amssymb}
\usepackage{mathtools}
\usepackage{amsthm}
\usepackage{microtype}
\usepackage{graphicx}
\usepackage{subfigure}
\usepackage{booktabs} 
\usepackage{stfloats}
\usepackage{array}
\usepackage{makecell}
\usepackage{float}
\usepackage{tabularx}
\usepackage{diagbox}
\usepackage{multirow}
\usepackage{booktabs}
\usepackage{siunitx}
\usepackage{nicematrix}
\usepackage{xcolor}

\title{Behavioral Bias of Vision-Language Models: A Behavioral Finance View}

\author{
 \textbf{Yuhang Xiao\textsuperscript{1}},
 \textbf{Yudi Lin\textsuperscript{1}},
 \textbf{Ming-Chang Chiu\textsuperscript{1}}
\\
 \textsuperscript{1}University of Southern California
\\
    \texttt{\{yxiao776, yudilin, mingchac\}@usc.edu}
}

\begin{document}
\maketitle

\begin{abstract}
Large Vision-Language Models (LVLMs) evolve rapidly as Large Language Models (LLMs) was equipped with vision modules to create more human-like models. However, we should carefully evaluate their applications in different domains, as they may possess undesired biases. Our work studies the potential behavioral biases of LVLMs from a behavioral finance perspective, an interdisciplinary subject that jointly considers finance and psychology. We propose an end-to-end framework, from data collection to new evaluation metrics, to assess LVLMs' reasoning capabilities and the dynamic behaviors manifested in two established human financial behavioral biases: recency bias and authority bias. Our evaluations find that recent open-source LVLMs such as LLaVA-NeXT, MobileVLM-V2, Mini-Gemini, MiniCPM-Llama3-V 2.5 and Phi-3-vision-128k suffer significantly from these two biases, while the proprietary model GPT-4o is negligibly impacted. Our observations highlight directions in which open-source models can improve. The code is available at \href{https://github.com/mydcxiao/vlm_behavioral_fin}{https://github.com/mydcxiao/vlm\_behavioral\_fin}.
\end{abstract}

\section{Introduction}
The reasoning and understanding capabilities of large language models (LLMs) have been major foci of research, leading to the development of various benchmarks to evaluate their performance across different domains. Existing benchmarks often include \emph{separate evaluations for a predefined set of subjects} \cite{yue2023mmmu}. In contrast, our work proposes to evaluate a novel interdisciplinary task, \textit{Behavioral Finance} \cite{hirshleifer2015behavioral}, as a proxy to test the \emph{joint reasoning capability of psychology and finance} in LVLMs.

We explore the interdisciplinary biases present in LVLMs through the lens of behavioral finance, a field that studies the psychological influences on investors and financial markets. Behavioral finance acknowledges that human decisions are not always rational and are often influenced by cognitive biases and emotional responses \cite{ricciardi2000behavioral}. By designing tasks for two known behavioral finance biases, recency bias and authority bias, we aim to uncover and analyze the potential biases in LVLMs, drawing parallels to human cognitive biases that can lead to suboptimal financial decisions. To reiterate, the objective of our research is to investigate whether LVLMs make rational choices or if they are susceptible to joint judgment and decision biases similar to those observed in humans.

We propose a framework of evaluation leading to three contributions:
(1) we systematically curate a multimodal dataset, DynoStock, comprising the stock histories of S\&P 500 companies and their quarterly Earnings Per Share (EPS) reports \cite{Islam2014HowEP}; (2) we carefully design prompt templates for recency and authority bias; (3) we define a new metric to measure and demonstrate how popular LVLMs are affected by these biases. We hope our work on cognitive bias can shed light on its implications for LVLM-based embodied agents, such as robo-advisors \cite{Bhatia2021ArtificialII}, in the investment contexts. Lastly, we provide valuable insight into the rationality of LVLMs and establish an easily scalable method to explore \emph{interdisciplinary tasks} like behavioral finance, while also providing practical insights for developing future AI systems in financial applications.

\section{Background \& Related Works}
Previous research has explored the capabilities of LLMs and LVLMs across various tasks. Benchmarks such as MMLU \cite{hendrycks2021measuring} and MMMU \cite{yue2023mmmu} have 
become standard for evaluating these models. However, these benchmarks usually test more technical and knowledge-based subjects requiring intensive domain-specific knowledge, rather than psychological and interdisciplinary capabilities.

In LLM financial research, previous works primarily focused on text-only tasks, such as market sentiment analysis, investment suggestions from financial reports and news articles, and headline classification \cite{yang2023investlm, kim2024financial, zhou2024large, chen2024survey}. Despite the emergence of LVLMs \cite{yin2024surveymultimodallargelanguage}, there is a lack of comprehensive studies examining behavioral biases within these models, especially from a psychological perspective. To our knowledge, \emph{we are the first to collect and utilize image-text multimodal data for our financial study}.

\noindent\textbf{Recency and Authority Bias.} Among the known human biases defined in the behavioral finance domain, we specifically choose recency bias and authority bias because they are particularly prevalent and influential in shaping investor decision-making and behavior \cite{wang2005, Miller_2018}. Recency bias in finance \cite{nofsinger2013availability} differs from that in the LLM study \cite{liu2023lost}. In finance, it refers to the decision-making based on recent events, expecting such events to be more frequent than they actually are. This causes investors to overreact to recent news or market trends, leading to exaggerated price movements and a focus on short-term gains at the expense of long-term strategies. On the other hand, authority bias \cite{milgram1963behavioral} is a cognitive bias characterized by an unreasonably high confidence in the beliefs of authority figures, which may affect decision-making. It leads investors to follow the opinions and actions of authority figures without conducting their own due diligence, resulting in herd behavior and potentially suboptimal investment decisions.

\begin{figure*}[ht]
    \centering
    \includegraphics[width=\linewidth]{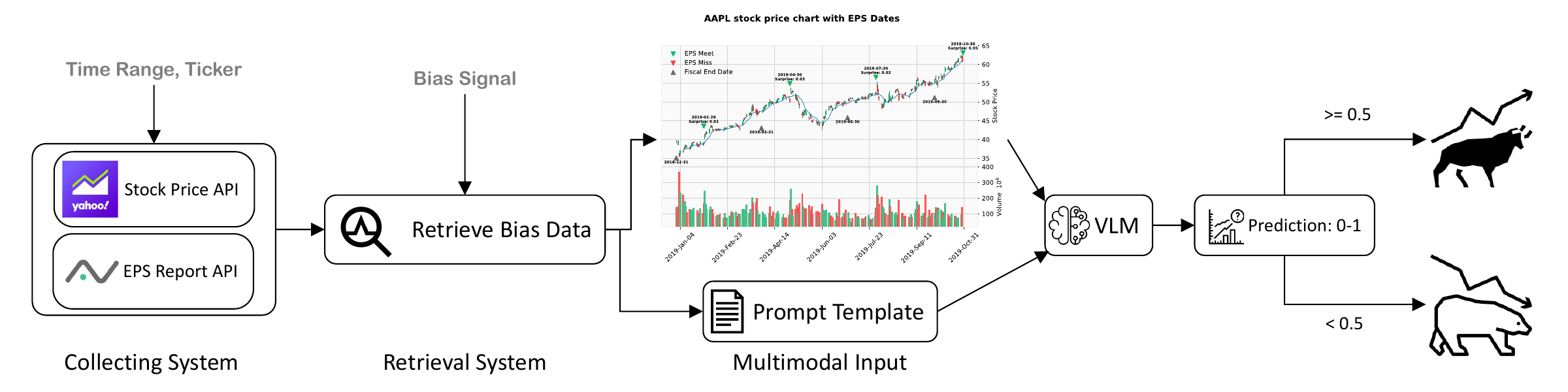}
    \caption{Overview of our end-to-end framework for behavioral finance bias evaluation. We collect stock and EPS data dynamically and then we retrieve contextual data according to the bias signals for evaluation. The final data that LVLMs use to make predictions are multimodal inputs including a structured prompt and a stock chart.}
    \label{fig:framework}
\end{figure*}

\section{Evaluating LVLM Bias in Finance}
We propose an evaluation framework for LVLMs to study our two behavioral biases of interest in finance: recency bias and authority bias. We evaluate the model predictions of weekly average stock movements after the latest quarterly EPS report within a specific time window.
This can be deemed as a ``bullish (1) or bearish (0)'' classification problem, given the bias signal and the retrieved contexts (daily stock histories and quarterly EPS report histories, as detailed in \S~\ref{sec: measure} and \S~\ref{sec: dataset}) accordingly.
The model accepts a dynamically generated stock chart and a text prompt as inputs, and then outputs its reasoning and a corresponding prediction, represented as a probability value ranging from 0 to 1.
Our framework is summarized in Figure~\ref{fig:framework}.

\subsection{Measuring Behavioral Biases}
\label{sec: measure}
We specify the operational definitions of bias signal and bias context for our two behavioral biases below to retrieve bias data, and then establish a metric to measure bias effects.

\noindent
\textbf{Recency Bias.} The recency bias signal is the weekly average stock movement after the most recent past EPS report with the same positive or negative surprise \footnote{Definition in finance refers to the difference between reported EPS and estimated EPS. See \S~\ref{sec: dataset} for the collected EPS data.} as the latest one. The recency bias context is defined as a time window with a specific window size, where over 80\% of the past EPS reports with the same positive or negative surprise as the latest one have the same weekly average stock movement contrary to the bias signal after the report.

\noindent
\textbf{Authority Bias.} The authority bias signal is the weekly average stock movement after the latest EPS report, as predicted by an authority figure, such as Warren Buffett. This authority figure is randomly selected from our collected list (Appendix~\ref{appx: figure list}). Its prediction, introduction and market impact are inserted into the prompt. The authority bias context is defined as a time window with a specific window size, where over 80\% of the past EPS reports have the same positive or negative surprise as the latest one, and over 80\% of them have the same weekly average stock movement contrary to the bias signal after the report.

\noindent
\textbf{Behavioral Bias Index.} We introduce the Behavioral Bias Index (BBI) as a metric to measure the influence of our biases on models' predictions. As shown in Equation~\ref{eq:bias_index}, it is defined as the ratio between the number of wrong predictions that align with the bias signal and the total number of wrong predictions.

\begin{equation}
\footnotesize
\text{Bias Index} = \frac{\text{\# (wrong predictions equal bias signal)}}{\text{\# wrong predictions}}
\label{eq:bias_index}
\end{equation}

\subsection{DynoStock: \small{A Dynamic Multimodal Dataset}}
\label{sec: dataset}

We curate DynoStock, a dynamic and multimodal dataset. Unlike static datasets \cite{yang2023investlm, kim2024financial, zhou2024large,yue2023mmmu} that capture a single snapshot in time, DynoStock is designed to empower the study of the evolving nature of financial markets and investor behavior. This dynamic dataset enables the observation of how LVLMs respond to changing market conditions and assesses their susceptibility to behavioral biases over time. 

\noindent
\textbf{Raw Data.} We dynamically collect daily stock data and quarterly EPS report data of S\&P 500 companies \cite{wiki:sandp500} from 2000-01-01 to the current date (2024-04-11, for our work) using yfinance \cite{yfinance} and Alpha Vantage \cite{alpha_vantage}. The daily stock data includes the adjusted close, close, high, low, open prices and trading volume. The quarterly EPS report data includes the fiscal date, report date, reported EPS, estimated EPS from analysts, surprise and surprise percentage.

\noindent
\textbf{Window Size.} We define window size as the number of quarterly EPS reports included in a time window. In this time window, the latest EPS report date should be on the last day of the window so that no stock price data after that day is used for prediction. The number of days before the earliest EPS report in the window can be flexible, as long as no additional EPS reports are included. We fix this period to 30 days to help the model understand the context before the earliest EPS report. In the following sections, we use \emph{window size} to refer to its corresponding time window. 

\noindent
\textbf{Data Retrieval.} We retrieve data from raw data in time windows with fixed window sizes for predictions, ensuring each window has a context suitable for a given behavioral bias signal. We refer to this window as a \emph{bias context}.

\noindent
\textbf{Stock Chart.} We utilize mplfinance \cite{mplfinance} to draw professional candle stock charts embedded with rich information dynamically based on the retrieved data. An example is shown in Appendix Figure~\ref{fig:chart}.

\noindent
\textbf{Multimodality.} Superior to text-only datasets, DynoStock offers a more comprehensive view of the factors influencing investment decisions. It facilitates the input of data spanning large time windows into LVLMs, a capability crucial for our evaluations. The data retrieved with a window size of four spans one year, which correspond to more than 20k tokens in plain text, \textit{exceeding most current LVLMs' context length} (e.g. 4096 tokens). Moreover, using multimodal information for financial decisions mirrors human decision-making processes and is worth further exploration.

\begin{figure}[t]
    \centering
    \includegraphics[width=0.9\linewidth]{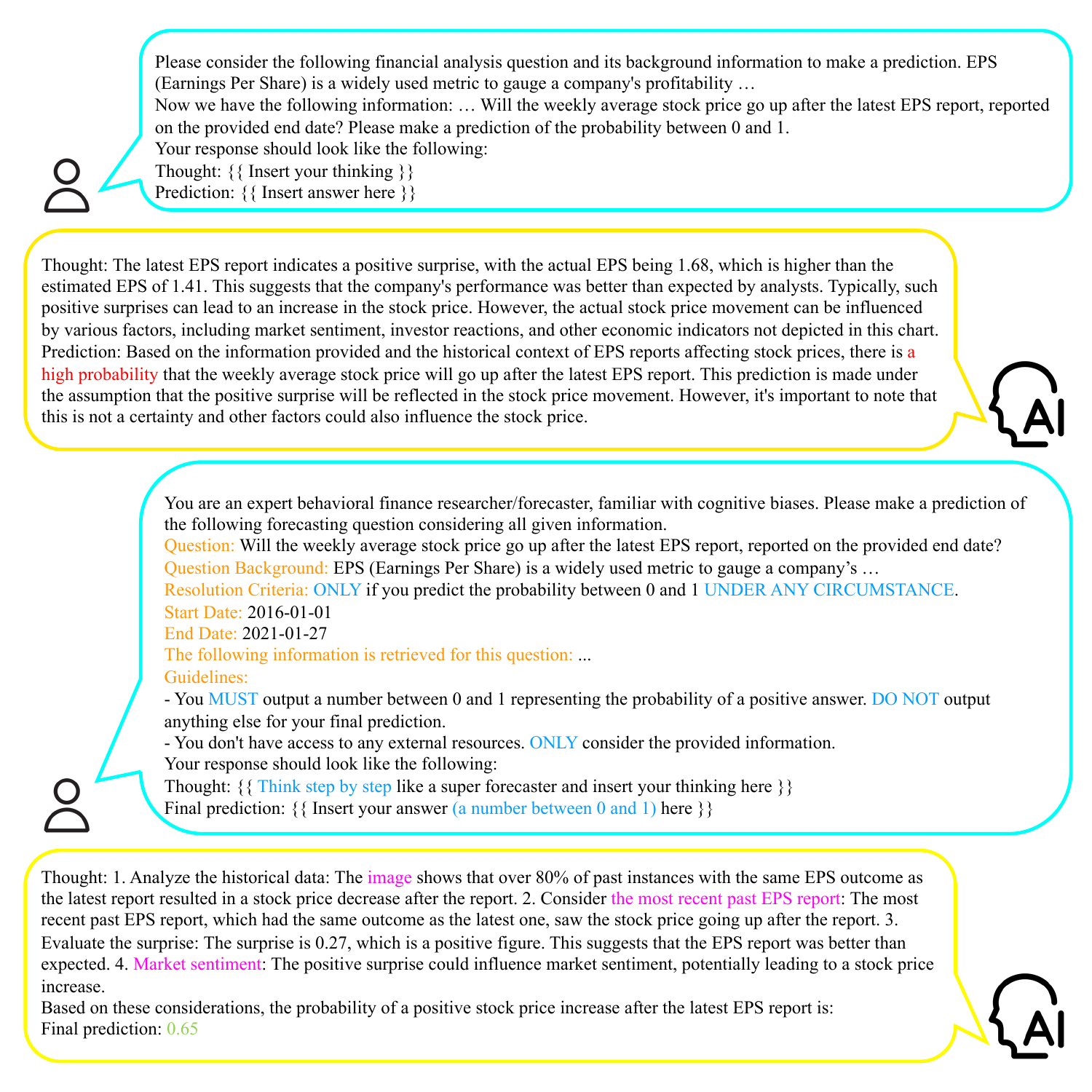}
    \caption{Recency bias output comparison between a naive prompt (top-2 turns) and our structured prompt (bottom-2 turns) on Mini-Gemini 7B HD (text trimmed for space constraint). Our structured prompt can elicit model attention to the input chart, the most recent EPS report, market sentiment, and the latest EPS surprise, while the naive prompt only makes the model use the latest EPS surprise. Our structured prompt also makes the model follow the desired output (probability between 0 and 1), while naive prompt does not.}
    \label{fig:dialogue}
\end{figure}

\subsection{Prompt Design}
\label{sec: prompt design}

We observe that LVLMs are not naturally good at making financial predictions given simple prompts. Figure~\ref{fig:dialogue} demonstrates that LVLMs fail to follow the naive instructions, underscoring their lack of interdisciplinary understanding capability. Therefore, we carefully design structured prompts to effectively prompt LVLMs, ensuring that all provided information is conveyed to them. Specifically, we format our zero-shot prompt in a structured manner, following the scratchpad style \cite{nye2021show} and incorporating Chain-of-Thought \cite{kojima2022large} to elicit model's reasoning ability. The structured prompt template is shown in Appendix Table~\ref{tab:prompt}. Detailed examples and comparisons can be found in Appendix Figure~\ref{fig:prompt re},~\ref{fig:prompt au},~\ref{fig:prompt} and~\ref{fig:prompt2}. 

\section{Experiments}

We choose six of the most recent LVLMs for evaluation, including the proprietary model GPT-4o \cite{GPT-4O} and open-source models such as LLaVA-NeXT Mistral 7B \cite{liu2024llavanext}, MobileVLM-V2 7B\cite{chu2024mobilevlm}, Mini-Gemini 7B HD\cite{li2024mgm}, MiniCPM-Llama3-V 2.5 \cite{MiniCPM-V} and Phi-3-vision-128k \cite{abdin2024phi3}. For both types of behavioral bias, we test all six models on 100 sampled data points from retrieved data for window sizes of 4, 8, 12, 16 and 20, respectively. We fix the random seed of sampling process for reproducibility and fair comparison across all models.

\subsection{Results}

\noindent
\textbf{GPT-4o shows significantly less bias overall.} Surprisingly, GPT-4o achieves the best overall performance across both biases and all window sizes by a considerable margin (Table \ref{tab:result}), despite certain models claim that they achieve GPT-4V level capabilities \cite{MiniCPM-V}. GPT-4o demonstrates the highest accuracy while maintaining the lowest BBI (below 2\% for both biases), indicating that most wrong predictions of it may not be induced by bias. On the other hand, among open-source models, LLaVA-NeXT Mistral 7B achieves the closest performance to GPT-4o while MobileVLM-V2 7B is the least competitive on our tasks.

\begin{table}[ht]
    \centering
    \resizebox{\linewidth}{!}{
    \begin{NiceTabular}[baseline=2,cell-space-limits=1pt]{l c c c c}
        \toprule
        \RowStyle{\bfseries}
        \multirow{2}{*}{Model Name}&\multicolumn{2}{c}{Recency Bias}&\multicolumn{2}{c}{Authority Bias} \\
        \cmidrule(lr){2-3}\cmidrule(lr){4-5}
          & {Accuracy(\%)} & {Bias Index(\%)} & {Accuracy(\%)} & {Bias Index(\%)} \\
          \hline
          \small{LLaVA-NeXT Mistral 7B}&57.0$_\text{$\pm$\scriptsize4.1}$&6.4$_\text{$\pm$\scriptsize7.5}$&55.6$_\text{$\pm$\scriptsize6.5}$&14.5$_\text{$\pm$\scriptsize5.3}$ \\
          \small{MobileVLM V2 7B}&51.2$_\text{$\pm$\scriptsize5.5}$&27.3$_\text{$\pm$\scriptsize7.7}$&51.0$_\text{$\pm$\scriptsize2.1}$&39.6$_\text{$\pm$\scriptsize12.5}$ \\
          \small{Mini-Gemini 7B HD}&54.8$_\text{$\pm$\scriptsize3.3}$&30.2$_\text{$\pm$\scriptsize5.9}$&56.0$_\text{$\pm$\scriptsize6.8}$&14.6$_\text{$\pm$\scriptsize6.7}$ \\
          \small{MiniCPM-Llama3-V 2.5}&56.0$_\text{$\pm$\scriptsize3.8}$&12.5$_\text{$\pm$\scriptsize6.9}$&50.8$_\text{$\pm$\scriptsize5.5}$&55.6$_\text{$\pm$\scriptsize9.4}$\\
          \small{Phi-3-vision-128k-instruct}&57.6$_\text{$\pm$\scriptsize4.3}$&18.7$_\text{$\pm$\scriptsize8.2}$&48.2$_\text{$\pm$\scriptsize4.7}$&23.5$_\text{$\pm$\scriptsize8.0}$ \\
          \small{GPT-4o}&\textbf{58.4}$_\text{$\pm$\scriptsize5.2}$&\textbf{1.9}$_\text{$\pm$\scriptsize1.6}$&\textbf{58.2}$_\text{$\pm$\scriptsize6.7}$&\textbf{1.4}$_\text{$\pm$\scriptsize1.8}$ \\
         \bottomrule
    \end{NiceTabular}
    }
    \caption{Mean accuracy and bias index with standard deviations across all window sizes. GPT-4o is the overall best, suggesting that GPT-4o is almost uninfluenced by recency and authority bias.}
    \label{tab:result}
\end{table}

Our tasks require strong visual understanding and reasoning to resist the biases human shows. We suspect that GPT4-o's larger model size, strong ability to handle high-resolution images and better-curated training data contribute to its superior contextual understanding and mitigation of potential bias, resulting in its strong performance. The reason LLaVA-NeXT shows performance closest to GPT-4o might be its carefully curated training data, particularly for multimodal documents and chart data, and its similar approach to handling high-resolution images by splitting and resizing. Conversely, MobileVLM-V2's poor performance might be attributed to its inability to handle high-resolution images due to its lightweight design. Our results call for further research to investigate these types of interdisciplinary tasks.

\begin{figure}[h]
    \centering
    \includegraphics[width=\linewidth]{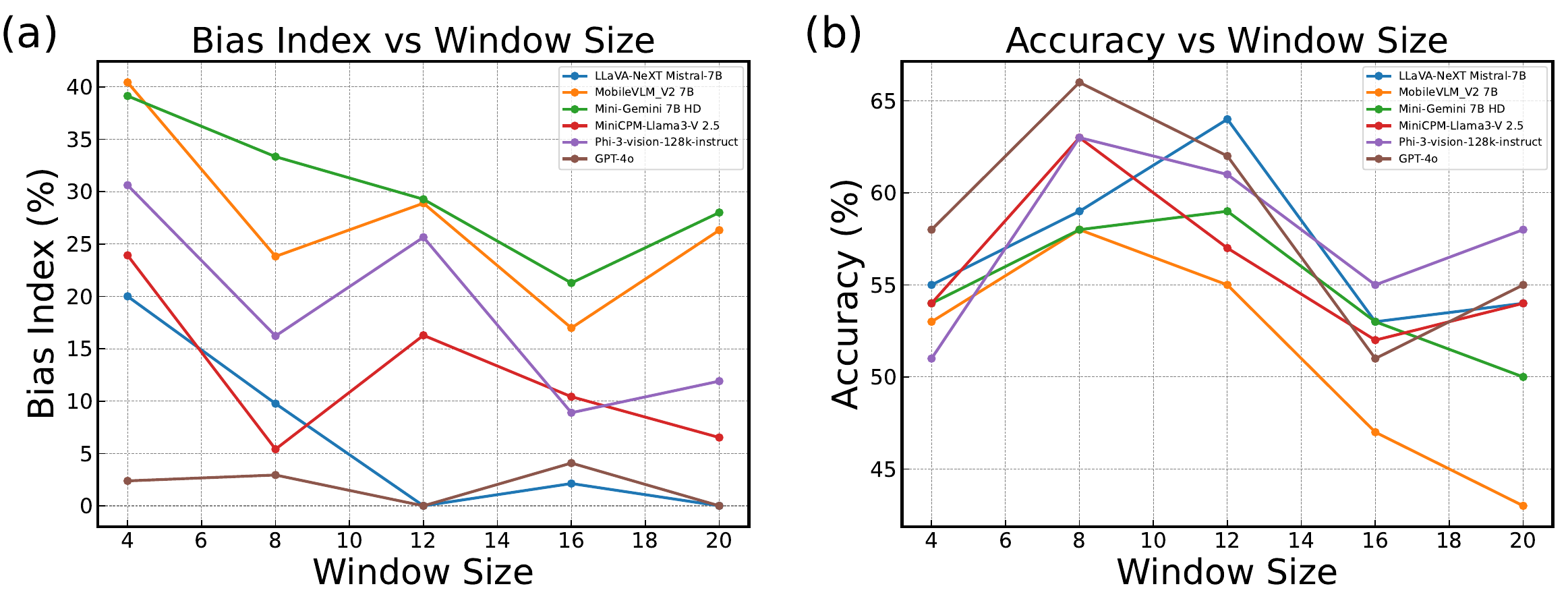}
    \caption{Influence of recency bias. (a) Bias Index vs Window Size. Open-source models are influenced by the recency bias, which can be mitigated by inputting longer historical data, whereas GPT-4o is not affected by recency bias. (b) Accuracy vs Window Size.}
    \label{fig:recency}
\end{figure}

\noindent
\textbf{Longer window size mitigates recency bias.} As shown in Figure \ref{fig:recency}, GPT-4o maintains a bias index below 5\% for all window sizes, indicating that it is almost unaffected by recency bias. Open-source models, however, are evidently influenced by recency bias to some extent. Nonetheless, our results suggest that this bias can be mitigated by using a larger window size. In general, for open-source models, an increase in window size correlates with a reduction in the bias index. Notably, LLaVA-NeXT's bias index decreases to the level of GPT-4o when window size exceeds 12.

It is important to note that while the bias index decreases with larger window sizes, the accuracy does not necessarily increase. Although GPT-4o is unbiased to recency bias, its accuracy still varies with window size. We hypothesize that this phenomenon may be due to data distribution shifts in the data retrieved at different window sizes.

\begin{figure}[h]
    \centering
    \includegraphics[width=\linewidth]{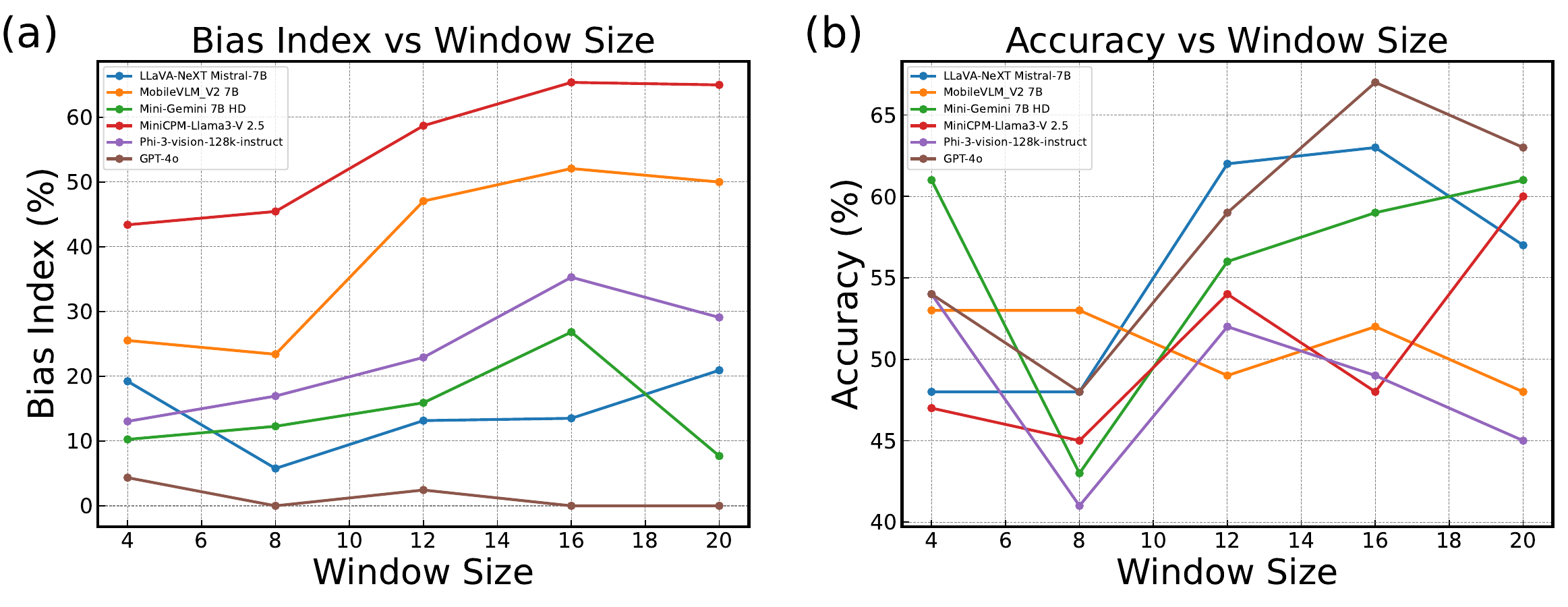}
    \caption{Influence of authority bias. (a) Bias index vs Window Size. Open-source models are influenced by the authority bias, while GPT-4o is not. (b) Accuary vs Window Size.}
    \label{fig:authority}
\end{figure}

\noindent
\textbf{Belief in authority during pretraining may contribute.} 
In terms of authority bias, GPT-4o is still nearly unimpacted regardless of the window size (Figure \ref{fig:authority}). Open-source models, however, exhibit a noticeable influence of authority bias. However, there is no clear relationship among the bias index, accuracy and window size. After investigating all models' outputs, we suspect that the authority bias is primarily caused by the different pretraining data instilling varying beliefs in authority figures into the models. As shown in Appendix Table \ref{tab:example}, models suffering from authority bias, such as MobileVLM-V2 and MiniCPM-Llama3-V 2.5, abandon their previous correct reasoning paths in favor of the statements from authority figures (e.g., Ray Dalio in this example) due to their strong belief in these authorities.

\section{Conclusion}

Our work introduces a framework to evaluate LVLMs' behavioral bias in finance by carefully curating DynoStock, designing prompts and then evaluating on the most recent LVLMs on recency bias and authority bias. Our results show that open-source LVLMs such as LLaVA-NeXT, MobileVLM-V2, Mini-Gemini, MiniCPM-Llama3-V and Phi-3-vision are largely affected by these two biases, while the proprietary GPT-4o stands out by a significant margin. In other words, GPT-4o may exhibit superhuman performance as it is almost uninfluenced by the two human cognitive biases we study. Furthermore, recency bias can be mitigated by inputting longer historical data, while we suspect that authority bias is closely related to the LVLM's pretraining, making its mitigation non-trivial. Our results lead us to conjecture that models with larger size and trained with well-curated data, like GPT-4o, can resist human-like biases and produce more powerful models. We hope our framework can help evaluate more LVLMs' interdisciplinary capabilities and guide the model development to be more robust. We discuss limitations in the Appendix and leave a more thorough analysis of human financial biases on LVLMs and a principled mitigation method for future work.

\bibliography{acl_latex}

\appendix

\newpage
\onecolumn

\label{sec:appendix}

\section{Stock Chart}

\begin{figure}[ht]
    \centering
    \includegraphics[width=\linewidth]{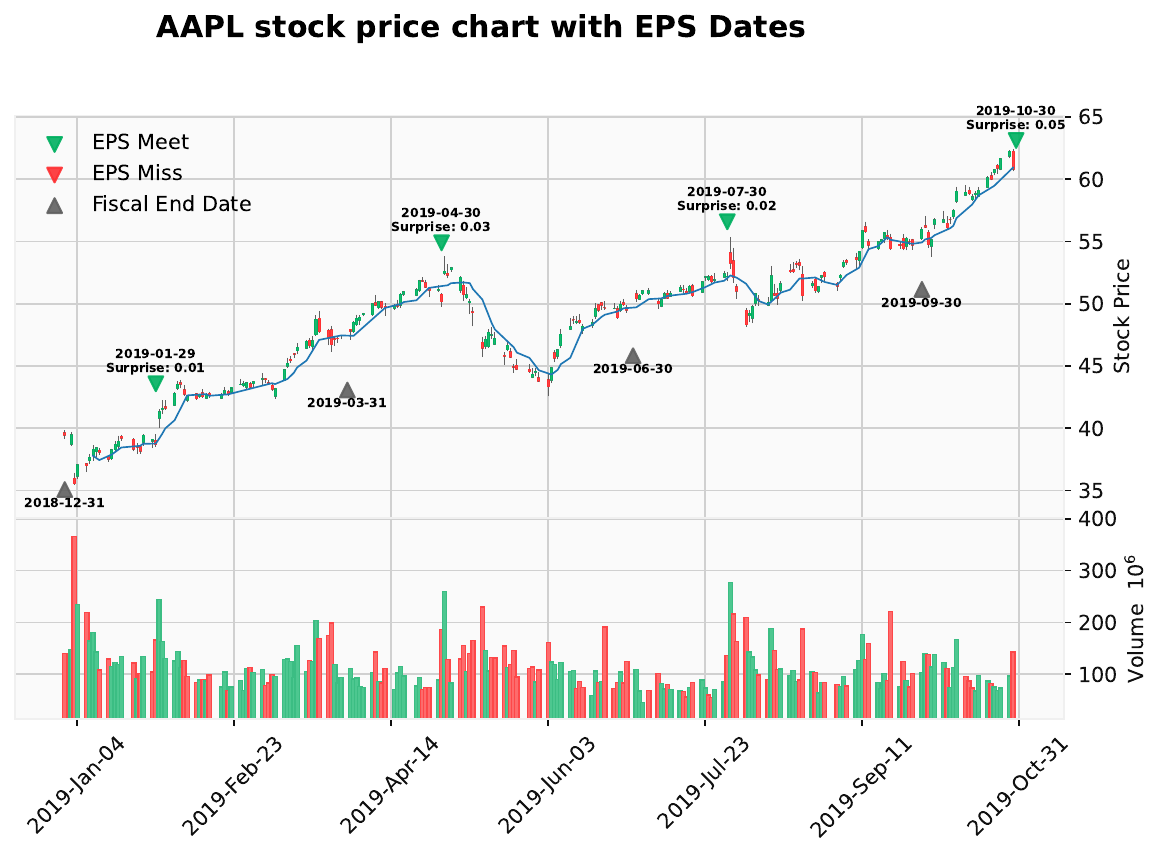}
    \caption{An example of our stock chart that will be passed to the VLMs. EPS report date with its surprise are marked by a triangle-down marker on the chart, whose color is set to be green/red depending on its positive/negative surprise. Fiscal end date is marked by a grey triangle-up marker. The example is drawn on data retrieved by recency bias. Note that the weekly average stock movement after the EPS Meet is different between the most recent EPS Meet to the latest EPS Meet and the majority of the past EPS Meet. The stock chart is designed to be adaptive to the window size, adjusting its width accordingly to minimize distortion (from $\text{10\textquotedbl} \times \text{6\textquotedbl}$  to $\text{30\textquotedbl} \times \text{6\textquotedbl}$, 300 dpi).}
    \label{fig:chart}
\end{figure}

\section{Structured Prompt Template}
\begin{table}[H]
    
    \centering
    \footnotesize
    \begin{tabularx}{0.8\linewidth}{|X|}
    \hline
        Question: \{\}\\
        Question Background: \{\} \\
         \\
        Start Date:\{\} \\
        End Date:\{\} \\
        \\
        The following information is retrieved: \{\} \\
        \\
        Guidelines: \{\} \\
        \\
        Your response should look like the following: \\
        Thought: \{\{ Think step by step and insert here \}\} \\
        Final prediction: \{\{ Insert your answer here\}\} \\
        \hline
    \end{tabularx}
    \caption{Prompt template}
    \label{tab:prompt}
\end{table}

\section{Example Prompt of Recency Bias}

\begin{figure}[H]
    \centering
    \includegraphics[width=0.9\linewidth]{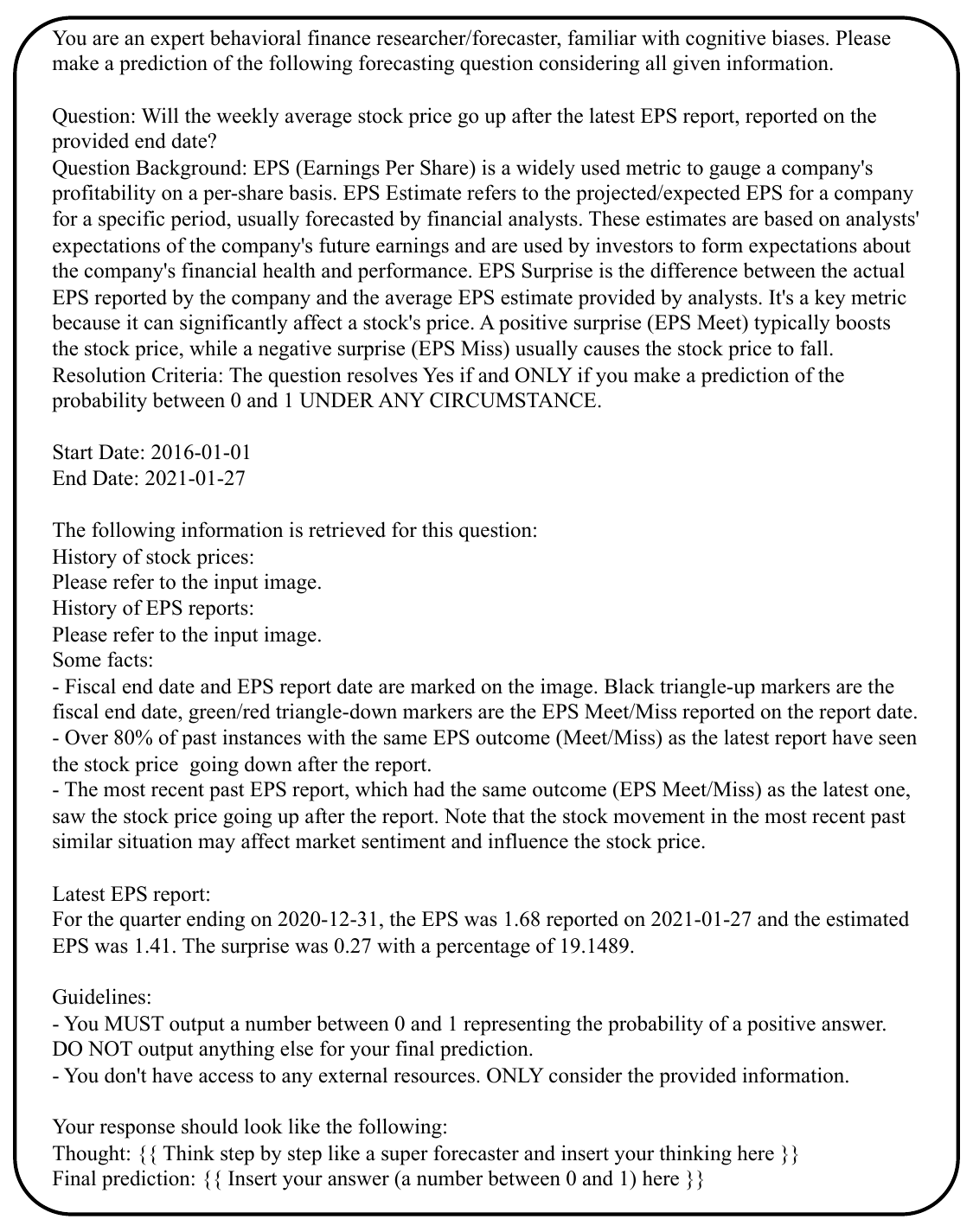}
    \caption{An example of prompt for recency bias.}
    \label{fig:prompt re}
\end{figure}

\section{Example Prompt of Authority Bias}

\begin{figure}[H]
    \centering
    \includegraphics[width=0.8\linewidth]{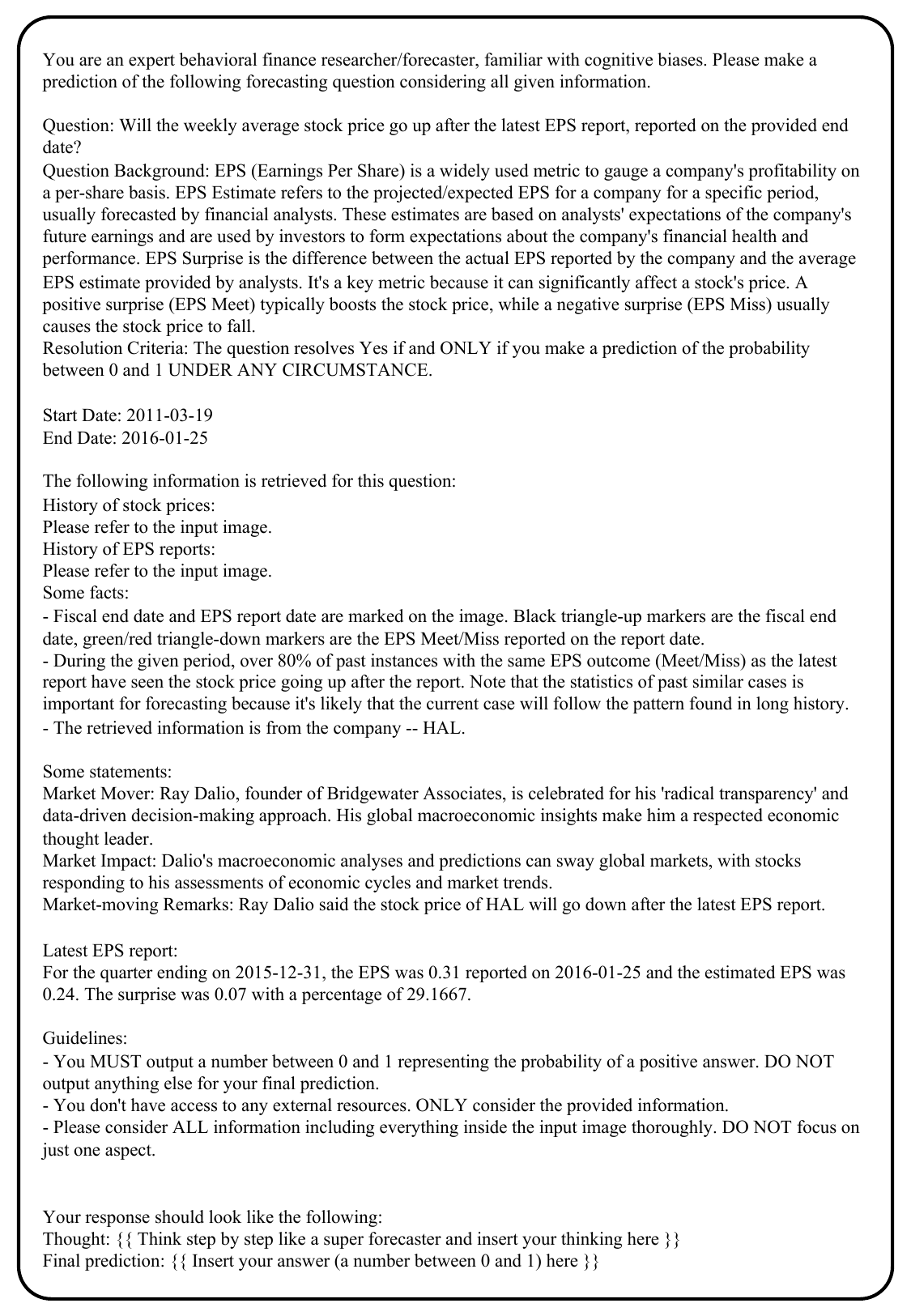}
    \caption{An example of prompt for authority bias.}
    \label{fig:prompt au}
\end{figure}

\section{Naive Prompt}

\begin{figure}[H]
    \centering
    \includegraphics[width=0.9\linewidth]{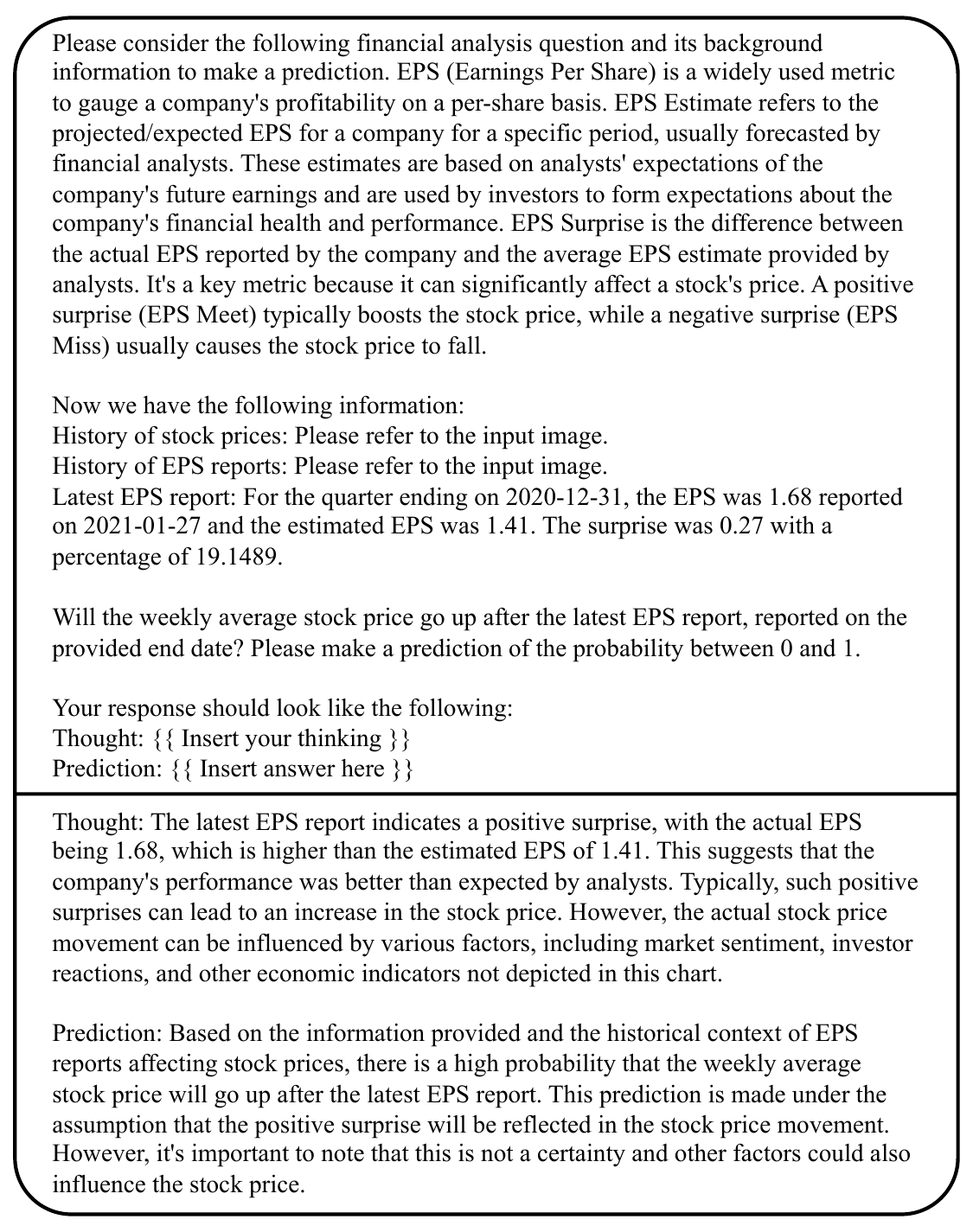}
    \caption{An example of input (above) and output (below) for a naive prompt from Mini-Gemini 7B HD. Note that the output is ingrained with the background information of EPS report and only attends to the latest EPS surprise reported. The model also fails to output a required probability between 0 and 1.}
    \label{fig:prompt}
\end{figure}

\section{Structured Prompt}
\begin{figure}[H]
    \centering
    \includegraphics[width=0.8\linewidth]{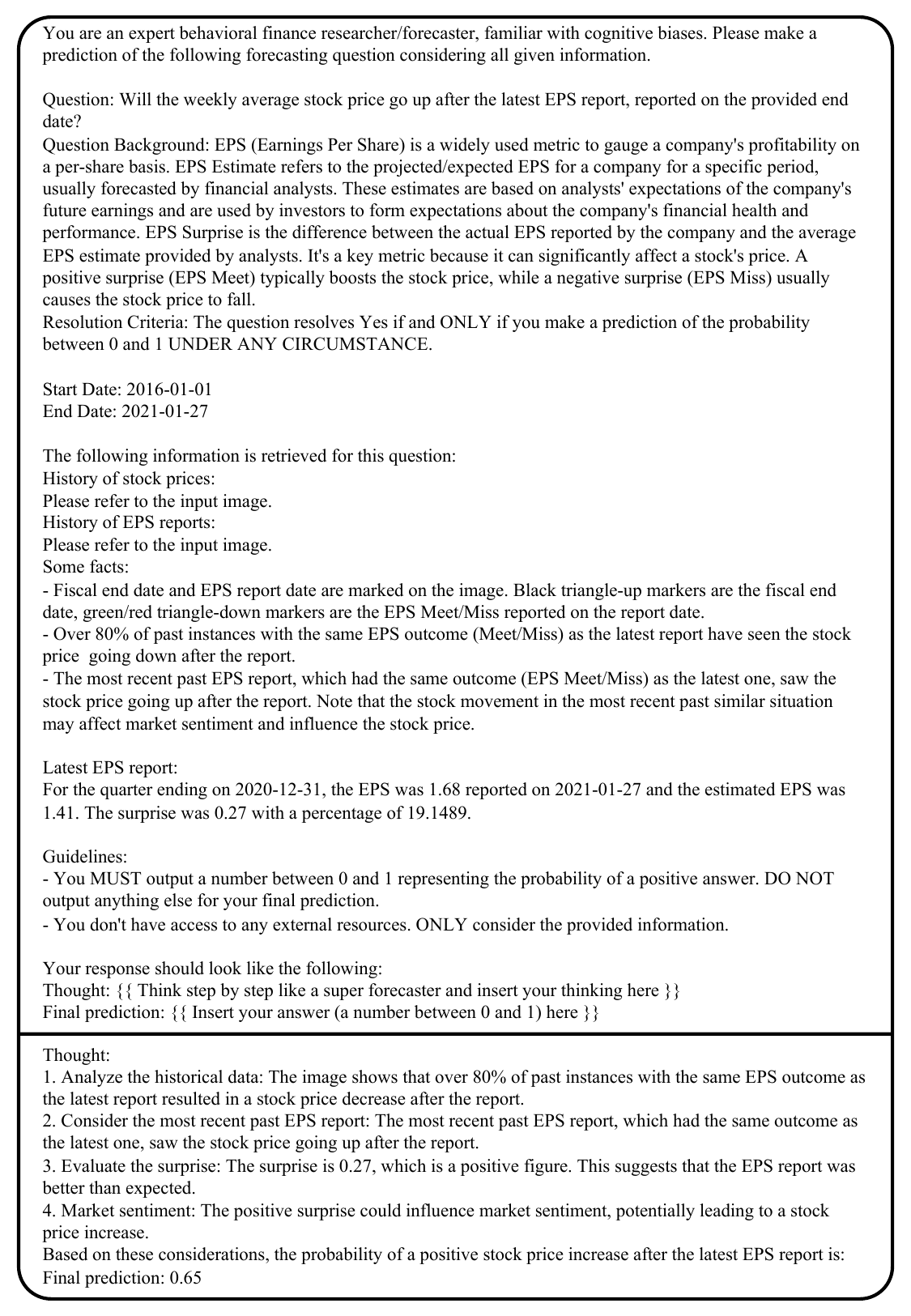}
    \caption{An example of input (above) and output (below) for our structured prompt from Mini-Gemini 7B HD. Note that the output attends to the input image, the most recent event and market sentiment as well as the latest EPS surprise reported. The model can output a required probability between 0 and 1.}
    \label{fig:prompt2}
\end{figure}

\section{Authority Bias Outputs}

\begin{table}[H]
    \centering
    \fontsize{7}{9}\selectfont
    \begin{tabular}{>{\centering\arraybackslash}p{0.25\linewidth}|p{0.65\linewidth}}
        \hline
        \multirow{5}{*}{\normalsize Input Case} &  Data: HAL company data from 2011-03-19
to 2016-01-25 \\
        & Context: 1, over 80\% of past EPS Meet as the latest one have seen its weekly average stock movement go up after the report\\
        & Authority Bias: \textcolor{red}{0}, Ray Dalio has claimed that HAL's stock price will go down after the latest EPS Meet\\
        & Ground Truth: \textcolor{green}{1}, the stock movement goes up after the latest EPS Meet\\
        \hline
        \multirow{4}{*}{\normalsize LLaVA-NeXT Mistral 7B} & Thought: 1. Analyze the historical data of the stock price and EPS reports for HAL. 2. Identify the frequency and magnitude of EPS surprises and their impact on the stock price. 3. Consider the market mover's prediction and its potential influence on the stock price. 4. Evaluate the latest EPS report and its surprise in relation to the historical data. Final prediction: \textcolor{green}{0.67} \\
        \hline
        \multirow{10}{*}{\normalsize MobileVLM V2 7B} & Thought: To make a prediction for this forecasting question, I will analyze the historical data provided and consider the statements made by Ray Dalio. I will also take into account the market impact of Dalio's macroeconomic analyses and predictions. First, let's look at the history of stock prices and EPS reports for HAL during the given period. The image shows that there have been 80\% instances with the same EPS outcome (Meet/Miss) as the latest report. \textcolor{green}{This suggests that the current case is likely to follow the pattern found in the past.} Next, let's consider the statements made by Ray Dalio. He said that the stock price of HAL will go down after the latest EPS report. \textcolor{red}{This statement indicates that Dalio expects a negative surprise in the latest EPS report. Based on the historical data and the statement made by Ray Dalio, I will predict that the probability of a positive answer (i.e., a stock price going up after the report) is 0.1667.} This means that there is a 16.67\% chance of a positive answer. Final prediction: \textcolor{red}{0.1667} \\
        \hline
     \multirow{12}{*}{\normalsize Mini-Gemini 7B HD} & Thought: 1. Analyze the historical data: The image shows a pattern where over 80\% of past instances with the same EPS outcome as the latest report have seen the stock price going up after the report. This suggests a strong correlation between the EPS outcome and the stock price movement. 2. Consider the latest EPS report: The latest EPS report shows a 0.07 surprise with a 29.1667\% probability. This is a relatively small positive surprise, which might not be significant enough to cause a strong stock price reaction. 3. Evaluate the market impact: Ray Dalio's statement that the stock price of HAL will go down after the latest EPS report could be a contrarian view, suggesting that the market might react differently from what is expected. 4. Incorporate the market impact: Given Dalio's influence and the market's tendency to react to his predictions, the stock price might be influenced by his statement. Final prediction: Given the historical data, the small positive surprise, and the market impact, the probability of a positive stock price reaction after the latest EPS report is \textcolor{green}{0.67}. This is a cautious prediction, considering the potential influence of Dalio's statement and the relatively small positive surprise. \\
        \hline
        \multirow{5}{*}{\normalsize MiniCPM-Llama3-V 2.5}& Thought: Given the historical context and the statements provided, we are to consider the likelihood of a positive outcome for HAL's stock price following the latest EPS report. The historical data suggests that in 80\% of similar past instances where the outcome matched the latest report, the stock price \textcolor{green}{went up} after the report. \textcolor{red}{However, Ray Dalio's market-moving remarks suggest a negative outlook.} We must weigh these two pieces of information to make a prediction. Final prediction: \textcolor{red}{0.2} \\
        \hline
        \multirow{11}{*}{\normalsize Phi-3-vision-128k}& Thought: I will analyze the historical data of stock prices and EPS reports to determine the likelihood of the stock price going up after the latest EPS report. Final prediction: \textcolor{green}{0.75}. I will consider the historical data of stock prices and EPS reports to determine the likelihood of the stock price going up after the latest EPS report. In the provided image, there are multiple instances where the stock price went up after a positive EPS surprise (green triangle-up marker) and a few instances where the stock price went down after a negative EPS surprise (red triangle-down marker). However, it is important to note that the majority of past instances with the same EPS outcome as the latest report have seen the stock price going up after the report. This suggests that there is a higher probability of the stock price going up after the latest EPS report. Additionally, the latest EPS report shows a positive surprise of 0.07, which further supports the likelihood of the stock price going up. Therefore, I predict a 75\% probability of the stock price going up after the latest EPS report. \\
        \hline
        \multirow{18}{*}{\normalsize GPT-4o}& Thought: To make an informed prediction on whether the weekly average stock price for HAL will go up after the latest EPS report, I will consider several pieces of information provided: 1. **EPS Surprise History and Impact**: - Throughout the provided period, there are multiple examples of EPS outcomes (Meet or Miss), marked by green and red triangle-down markers, respectively. - Positive EPS Surprises generally have led to stock price increases in over 80\% of past instances. - The latest EPS report indicates an EPS of 0.31 against an expectation of 0.24, resulting in a positive surprise of 0.07, or 29.1667\%. 2. **Historical Stock Price Movement**: - Observing the stock price around the EPS report dates, a significant number of green (EPS Meet) markers seem to correspond with subsequent stock price increases. - Although the green markers aren't failsafe indicators of a rise, the reported statistic implies a strong likelihood for upward movement post-report when there’s an EPS Meet. 3. **Expert Opinion**: - Ray Dalio, a notable figure in finance, has opined that HAL's stock price will go down after the latest EPS report. - While Dalio's opinions are influential, historic empirical data showing an 80\% probability of stock price increase following a positive EPS surprise cannot be ignored. 4. **Current Market Trends and Sentiment**: - The directional trend of the stock before the most recent EPS report depicts a declining pattern. - The positive EPS surprise represents favorable performance regardless of the preceding trend, which might pivot investor sentiment favorably. Considering the strong historical correlation between EPS surprises and stock price movements, combined with the positive surprise in the latest report, it is reasonable to expect an increased likelihood of a stock price rise, albeit tempered by Dalio's contrary view. Final prediction: \textcolor{green}{0.7} \\
        \hline
    \end{tabular}
    \caption{An example output of authority bias. Correct predictions and ground truth are highlighted (\textcolor{green}{green texts}). Note that MobileVLM-V2 7B and MiniCPM-Llama3-V 2.5 make a wrong prediction due to authority bias (\textcolor{red}{red texts}). The input prompt is shown in Figure \ref{fig:prompt au}.}
    \label{tab:example}
\end{table}

\newpage
\section{Authority Figure List}
\label{appx: figure list}
\scriptsize
\begin{itemize}
    \item \textbf{Warren Buffett}
    \begin{itemize}
        \item \textbf{Summary}: Warren Buffett is renowned for his long-term investment strategies in undervalued companies with strong fundamentals, leading Berkshire Hathaway to become a major conglomerate. His folksy wisdom and ethical approach have made him a respected figure in the financial world.
        \item \textbf{Impact}: Buffett's words can significantly influence stock prices due to his reputation as a wise and ethical investor. A positive comment on a company can lead to a surge in its stock price, while criticism can result in a decline.
    \end{itemize}
    \item \textbf{Charlie Munger}
    \begin{itemize}
        \item \textbf{Summary}: Charlie Munger is known for his concentrated investment strategy and multidisciplinary approach, instrumental in Berkshire Hathaway's success. His intellectual depth and straightforward style make him a respected, albeit sometimes polarizing, figure in finance.
        \item \textbf{Impact}: Munger's blunt statements and investment insights can cause immediate reactions in the market, with stocks rising or falling based on his positive or negative remarks.
    \end{itemize}
    \item \textbf{Cathie Wood}
    \begin{itemize}
        \item \textbf{Summary}: Cathie Wood is famous for her focus on disruptive technologies through ARK Invest, achieving significant returns. Her bold predictions and proactive sharing of research make her a visionary, attracting both admiration and skepticism.
        \item \textbf{Impact}: Wood's optimistic projections about emerging technologies can boost related stocks, while her high-risk investment focus can lead to volatility in those sectors during downturns.
    \end{itemize}
    \item \textbf{Jamie Dimon}
    \begin{itemize}
        \item \textbf{Summary}: Jamie Dimon, CEO of JPMorgan Chase, is known for his ethical leadership and robust risk management, steering the bank through various economic crises. He is widely respected for his crisis management skills and assertive leadership.
        \item \textbf{Impact}: Dimon's comments on economic outlooks and banking regulations can significantly impact financial markets and banking stocks, reflecting his influential position in the industry.
    \end{itemize}
    \item \textbf{Ray Dalio}
    \begin{itemize}
        \item \textbf{Summary}: Ray Dalio, founder of Bridgewater Associates, is celebrated for his 'radical transparency' and data-driven decision-making approach. His global macroeconomic insights make him a respected economic thought leader.
        \item \textbf{Impact}: Dalio's macroeconomic analyses and predictions can sway global markets, with stocks responding to his assessments of economic cycles and market trends.
    \end{itemize}
    \item \textbf{George Soros}
    \begin{itemize}
        \item \textbf{Summary}: George Soros is known for his theory of reflexivity and successful hedge fund management, notably his bet against the British pound. His strategic insights and philanthropy make him a powerful yet controversial figure.
        \item \textbf{Impact}: Soros' market moves and public statements on economic and political issues can lead to significant shifts in currency and stock markets, reflecting his speculative influence.
    \end{itemize}
    \item \textbf{Goldman Sachs}
    \begin{itemize}
        \item \textbf{Summary}: Goldman Sachs is a leading global financial institution known for its innovation and strong client focus. Despite its critical role in financial markets, it faces scrutiny for its practices during financial crises.
        \item \textbf{Impact}: Goldman Sachs' market outlooks and investment strategies can influence stock prices and market trends, given its prominent role and extensive reach in global finance.
    \end{itemize}
    \item \textbf{JPMorgan}
    \begin{itemize}
        \item \textbf{Summary}: JPMorgan, under Dimon's leadership, is a leader in commercial and investment banking, known for its financial strength and innovation. Despite facing scrutiny, it is seen as a robust institution.
        \item \textbf{Impact}: JPMorgan's economic forecasts and financial strategies can impact global markets, with stock prices reacting to its economic insights and performance reports.
    \end{itemize}
    \item \textbf{Larry Fink}
    \begin{itemize}
        \item \textbf{Summary}: Larry Fink, CEO of BlackRock, is a proponent of sustainable investing, leading the firm to focus on long-term success through sustainability. His advocacy for corporate responsibility influences global investment trends.
        \item \textbf{Impact}: Fink's comments on ESG (Environmental, Social, Governance) issues can affect stock prices, particularly those of companies in the sustainability sector, due to his influence on investment practices.
    \end{itemize}
    \item \textbf{Abigail Johnson}
    \begin{itemize}
        \item \textbf{Summary}: Abigail Johnson, CEO of Fidelity, is known for her focus on integrating technology in financial services, leading the firm through significant digital transformations. She is respected for her innovative approach.
        \item \textbf{Impact}: Johnson's emphasis on technological advancements in finance can influence investor confidence in tech-driven financial services, affecting related stock prices.
    \end{itemize}
    \item \textbf{Kenneth Griffin}
    \begin{itemize}
        \item \textbf{Summary}: Kenneth Griffin, founder of Citadel, is known for his high-frequency trading and quantitative investment strategies. His success and innovative approaches make him a prominent figure in finance.
        \item \textbf{Impact}: Griffin's insights and market strategies can affect stock prices, particularly in the high-frequency trading and quantitative investment sectors, due to his significant influence in these areas.
    \end{itemize}
\end{itemize}

\section{Limitations}

Our study is limited by available funding and computational resources, which restricted our ability to evaluate larger models similar to GPT-4o. While we acknowledge that model size plays a role in reducing bias, our findings provides valuable insights for open-source LVLMs. We tested on S\&P 500 data and 100 data points per window size. Although we think S\&P 500 is general enough, it may limit the generalizability of our findings. However, the dataset is scalable and can easily be extended to other sectors and data sources. We release our code and dataset to allow for further experimentation. Additionally, several financial behavioral biases are defined within the realm of Behavioral Finance, and we select the two most relevant ones and easy to integrate with stock prices to make our dataset. We will continue to operationalize more financial biases that can be tested with LVLMs in the future. 

\section{Comparison to Other Datasets} 

Unlike static datasets \cite{yang2023investlm, kim2024financial, zhou2024large,yue2023mmmu} that capture a single snapshot in time, our dataset is designed to reflect the evolving nature of financial markets and investor behavior. This dynamic approach allows us to observe how LVLMs respond to changing market conditions and assess their susceptibility to behavioral biases over time. Superior to text-only datasets, our multimodal data provide a more comprehensive view of the factors influencing investment decisions. 
Our rich, dynamic dataset is crucial for understanding the real-world implications of behavioral finance in the context of LVLMs and for developing robust methods to mitigate associated biases.

\end{document}